\documentclass[journal,twoside,web]{ieeecolor}
\usepackage{tmi}
\usepackage{cite}
\usepackage{color}
\usepackage{makecell}
\usepackage{multirow}
\usepackage{amsmath,amssymb,amsfonts}
\usepackage{hyperref}
\usepackage{algorithmic}
\usepackage{graphicx}
\usepackage{textcomp}
\def\BibTeX{{\rm B\kern-.05em{\sc i\kern-.025em b}\kern-.08em
    T\kern-.1667em\lower.7ex\hbox{E}\kern-.125emX}}
\begin{document}
\title{Cervical Glandular Cell Detection from Whole Slide Image with Out-Of-Distribution Data}
\author{Ziquan Wei, Shenghua Cheng, Jing Cai, Shaoqun Zeng, Xiuli Liu, and Zehua Wang,  
\thanks{This work is supported by the NSFC projects (grant 61721092), China Postdoctoral Science Foundation (grant 2021M701320) and the director fund of the WNLO.}
\thanks{Z. Wei and S. Cheng contributed equally to this work}
\thanks{Corresponding author: X. Liu and Z. Wang}
\thanks{Z. Wei, X. S. Cheng, S. Zeng , and X. Liu are with Collaborative Innovation Center for Biomedical Engineering, Wuhan National Laboratory for Optoelectronics-Huazhong University of Science and Technology, Wuhan, Hubei, China. They are also with Britton Chance Center and MOE Key Laboratory for Biomedical Photonics, School of Engineering Sciences, Huazhong University of Science and Technology, Wuhan, Hubei, China (email: wzquan142857@hust.edu.cn; chengshen@hust.edu.cn; sqzeng@mail.hust.edu.cn; xlliu@mail.hust.edu.cn).}
\thanks{J. Cai and Z. Wang are with Department of Obstetrics and Gynecology, Union Hospital, Tongji Medical College, Huazhong University of Science and Technology, Wuhan, China (e-mail: caijingmmm@hotmail.com; zehuawang@163.net).}}

\maketitle

\begin{abstract}
Cervical glandular cell (GC) detection is a key step in computer-aided diagnosis for cervical adenocarcinomas screening. It is challenging to accurately recognize GCs in cervical smears in which squamous cells are the major. Widely existing Out-Of-Distribution (OOD) data in the entire smear leads decreasing reliability of machine learning system for GC detection. Although, the State-Of-The-Art (SOTA) deep learning model can outperform pathologists in preselected regions of interest, the mass False Positive (FP) prediction with high probability is still unsolved when facing such gigapixel whole slide image. This paper proposed a novel PolarNet based on the morphological prior knowledge of GC trying to solve the FP problem via a self-attention mechanism in eight-neighbor. It estimates the polar orientation of nucleus of GC. As a plugin module, PolarNet can guide the deep feature and predicted confidence of general object detection models. In experiments, we discovered that general models based on four different frameworks can reject FP in small image set and increase the mean of average precision (mAP) by $\text{0.007}\sim\text{0.015}$ in average, where the highest exceeds the recent cervical cell detection model 0.037. By plugging PolarNet, the deployed C++ program improved by 8.8\% on accuracy of top-20 GC detection from external WSIs, while sacrificing 14.4 s of computational time. Code is available in \href{https://github.com/Chrisa142857/PolarNet-GCdet}{https://github.com/Chrisa142857/PolarNet-GCdet}.
\end{abstract}

\begin{IEEEkeywords}
Cervical Glandular Cell, Computer-aided Diagnosis, Deep Learning, Object Detection, Whole Slide Image Detection.
\end{IEEEkeywords}

\section{Introduction}
\label{sec:introduction}
\IEEEPARstart{D}{etecting} glandular cell (GC) from whole slide image (WSI) can aid pathologists to screen cervical adenocarcinomas (ADCA) and reduce the labour cost by computer-aided diagnosis (CAD) \cite{RN106, pirovano2021computer, li2019bayesian}. With the proper amount of annotated data, the well-trained deep learning model can provide top cells from WSI to help the screening, as a recommending system. The task is challenging, however, that squamous cell (SC) is the major in cervical smear. The sparseness of GC is severe. Otherwise, the morphology of GC and SC is similar while the basic shape of polar pattern of GC is distinct comparing with SC. Such characteristics are the direct hardship to the data-driven model performing accurately without the usage of the prior knowledge of GC basic shape in general models.

\begin{figure}[!t]
\centerline{\includegraphics[width=\columnwidth]{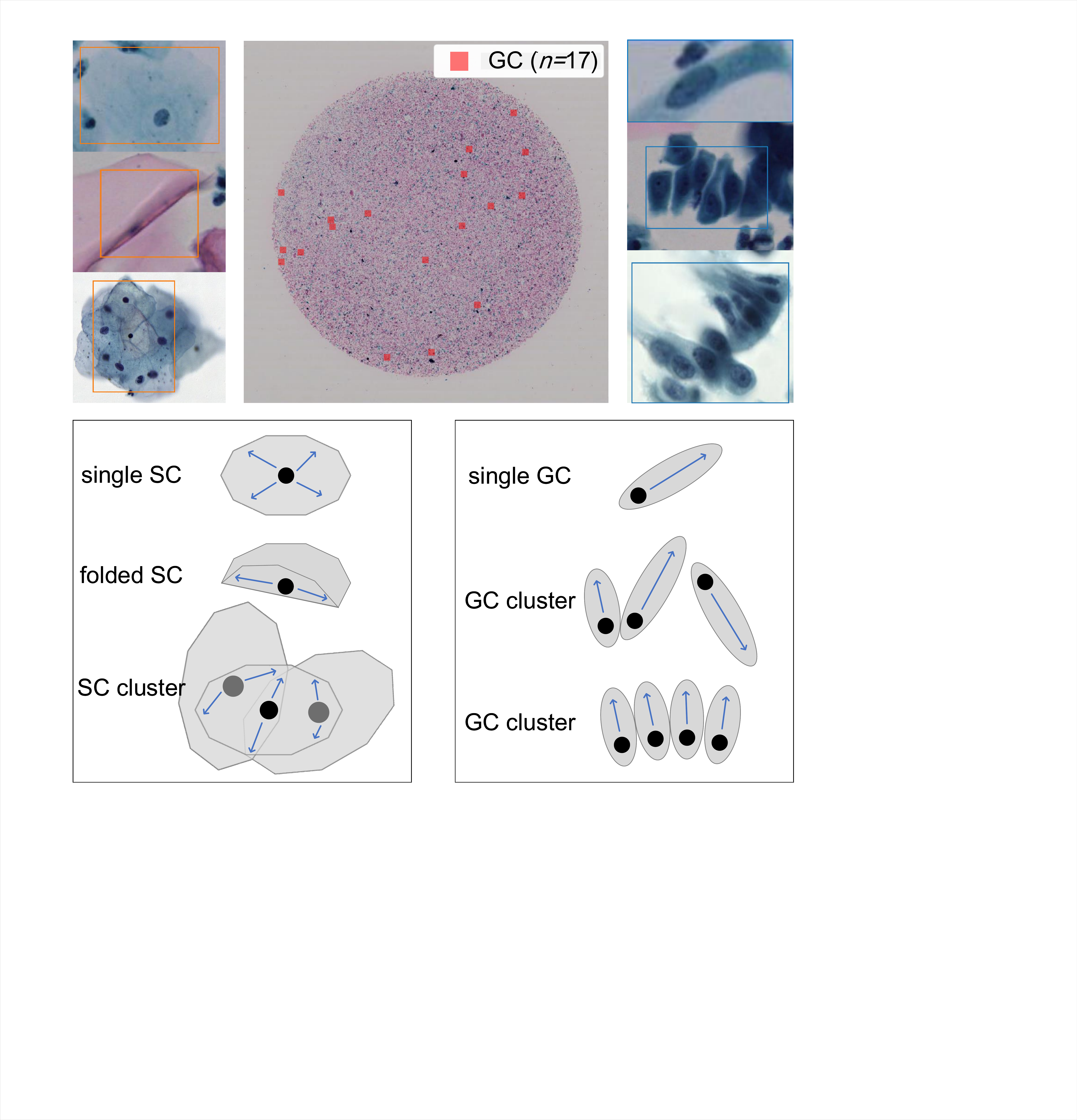}}
\caption{Upper figures: Glandular cell (GC) in cervical whole slide image is relatively sparse, where blue box denotes GC, and orange denotes squamous cell (SC). Below figures: Inconsistent polarity of SCs that are folded or clustered is in the left, and illustrations of polar pattern of GC are in the right, where blue arrow denotes polar orientation.}
\label{fig1}
\end{figure}

Although recognition cervical lesion in cytology has been developed since 2016 by using deep learning \cite{pouliakis2016artificial, van2016deep}, related works focus on the major object, SC, but rarely discuss GC, which is minor but important \cite{jiang2022deep}. They researched cervical cancer detection on small image sets with general deep learning methods, only few works mentioned the performance of model on external WSI set. Like Liang et al. \cite{RN162} tested the detection performance of both SC and GC 
with the deep learning model in only preseleted ROIs of external WSI. And Gupta et al. \cite{gupta2020region} only reported a rough detection results from external WSIs by their ROI prediction model. There is no focus on GC in cervical cytology and no usage of prior knowledge of GC in related works inspire that there is still room of improvement of GC detection using deep learning. The lack of evaluations of cell detection from WSI also leads related works are hardly deployed to the clinical CAD.

More vivid description of this problem is illustrated in Fig. \ref{fig1}, there are only 17 GCs existed in the WSI shown in the upper. This sparseness leads the computation of WSI will meet substantial out-of-distribution (OOD) data including artifacts, single SC, folded SC, and SC cluster. Even though their basic shape has obvious distinction comparing with GC as shown in the below of Fig. \ref{fig1}, general deep learning model is hard to distinguish. Those orange boxes are predicted by the state-of-the-art (SOTA) object detection model, YOLOX \cite{ge2021yolox}. Related works for cervical cell detection are designed based on more abstract knowledge, such as the attention and connection between different scales \cite{cao2021novel, wei2021efficient}, the mixed information of multiple resolutions \cite{cheng2021robust}, the prototype representation of multiple cell categories \cite{RN162} , and using time series information to enhance the cancer detection \cite{zhang2021quantitative}. Despite atypical GC is among their objects, their design concepts have not consider the basic shape of cervical cell. That is hardly helping to reject those false positives with wrong basic shape.

This paper proposed a new network to enhance GC detection with prior knowledge of basic shape of GC. We also provide a new experimental setting to show the practical performance of model. Firstly, a polar attention network (PolarNet) is designed to quantify the confusing morphology of GC, that is the basic shape of polar orientation. The network employs a novel self-attention mechanism in eight-neighbor that enables it to score the salience of polarity orientations of GC. Pseudo-GCs that are not significantly polar in the outer test set are controlled by such polarity role. The network is also a deep learning module that can be plugged into any general object detection model. Then, two additional test sets of small images from the external WSI containing different volume of OOD data are obtained by human and well-trained model, respectively. Furthermore, a elegant deployment of the proposal is completed and provided as a C++ program.

In GC detection experiments on both three small image sets, the proposed PolarNet showed effectiveness, and compared with 5 models including four different frameworks. With the usage of PolarNet, model can reject false positive (FP) and increase the mean of average precision (mAP) from 0.007 to 0.015 in average, where the highest exceeds the recent cervical cell detection model 0.037. The deployed program improved by 8.8\% on accuracy of top-20 GC detection from external WSIs ($n=110$), while sacrificing 14.4 s of computational time.

Briefly, this work proposed a cervical GC detection method to recognize GC from WSI, it consists of four contributions: 
\begin{itemize}
    \item In Section \ref{sec:datapreparation}, an OOD data set is provided using a well-trained YOLOX. The data is obtained from the severe FP predicted by the deep learning model, and is able to reflect the performance of GC detection from WSIs that are not fully annotated.
    \item In Section \ref{sec:method}, a novel eight-neighbor self-attention mechanism is proposed to quantify the polar orientation of GC and to construct the plugin network PolarNet. It can be used in general detection model whether it is single-stage or multi-stage to reject severe FP.
    \item In Section \ref{sec:exp}, the usage of PolarNet in four different frameworks both shown significance for GC detection. For clinical application, the proposal also performs effectiveness to recommend top-20 GCs from WSI.
    \item The proposed method is deployed as a C++ program. The computational cost of PolarNet is listed in detail. The program is released online.
\end{itemize}

\section{Related Works}
\label{sec:relatedworks}

\subsection{Cervical Cell Detection}
Earlier, the traditional framework was based on cytological definitions and completed cell classification by calculating the nuclear-cytoplasmic ratio of cervical cells, while Tareef et al. To improve the accuracy of subsequent classification tasks, their convolutional neural network (CNN) models surpass traditional machine learning algorithms in segmentation accuracy. But then, Zhang et al. \cite{RN61} believed that the unavoidable segmentation error would always lead to a decrease in the classification accuracy of abnormal cells, and they proposed the DeepPap model for the first time to directly apply the convolutional neural network to the classification task of cervical cells to avoid pre-segmentation processing, and performed well on two public datasets, Herlev 2005 \cite{sukumar2015computer} and HEMLBC \cite{RN59}, showing more than 98\% accuracy on both datasets, but their method still compares with traditional algorithms. more time consuming. Shanthi et al. \cite{RN110} explored in more detail the classification accuracy that can be obtained when using CNN for cervical cell classification, performing cell edge extraction, cell nucleus segmentation, or directly using the original image without segmentation, although their model is described in Herlev 2005. The accuracy rate is only 94\% $\sim$ 95\%, and it does not fully surpass the previous method, but it has been verified that it is the most effective to use the original image directly without segmentation. These early developments and conclusions using deep learning led to a perception of the potential of deep learning in CAD.

Later, with the development of general deep learning models, many general models such as VGG \cite{RN115}, GoogLeNet \cite{szegedy2015going}, ResNet \cite{he2016deep} and Inception \cite{szegedy2016rethinking} pre-trained and validated by the large general dataset ImageNet \cite{deng2009imagenet} appeared. Wait. Lin et al. \cite{RN111} proposed that these pre-trained models from general data can extract general morphological features, and they transferred these pre-trained models to the cervical cell classification task, using GoogLeNet in Herlev 2005 to obtain the highest accuracy of 94.5\%. In order to further improve the computational efficiency of CNN in cervical cell classification, Dong et al. \cite{RN153} proposed to combine a lightweight convolutional neural network with artificial features, by adapting the prior knowledge based on cytology definition to the Inception V3 model \cite{szegedy2016rethinking}, and finally achieved over 98\% accuracy on the public dataset Herlev 2005 \cite{sukumar2015computer}.

Because of the irregular distribution of cells on the slide in cervical cytology images, it is important to predict both the location and class of cells. Recently, Xiang et al. \cite{RN66}, Liang et al. \cite{RN159} proposed a cervical cell detection model earlier based on the single-stage object detection model YOLO (You Only Look Once) \cite{RN218}, and they stacked additional Inception V3, and FPN (Feature Pyramid Network) with content-aware function \cite{lin2017feature}, on their private 10-class cervical cell dataset, end-to-end localization, classification and classification of cervical cells were completed. Prediction size, and the average accuracy reached 63.4\%, although their image pixels are sufficient ($4000\times3000$), but the number of images is small ($n=12909$) is a defect. In order to solve the problem of model training when the amount of data is small, Liang et al. \cite{RN162} subsequently proposed a comparison detector based on a double-stage Faster RCNN. On small-scale ($n=7410$) datasets, the contrastive detectors can achieve significant improvements over the previous ones. The above are all cell images obtained by traditional production methods. 

In the work of Tan et al. \cite{RN168}, the authors firstly collected more than 16000 LBC (Liquid-Based Cytology) images as training and validation data. Compared with traditional cervical cytology images, the LBC image has a clearer background \cite{RN89}. They independently set 290 ROI (Region Of Interest) images in the external full slide as the test set to simulate real CAD process. They obtained decent accuracy after training a Faster RCNN model. The experiment of this deep learning model on external data provides preliminary feasibility support for CAD, although the amount of external data is still smaller than the training and validation data. 

Since the cervical cell detection task requires a larger field of view of image data than the earlier classification and segmentation tasks, and such public data sets are scarce, the above related object detection research is carried out on private data. More important, whether it is cell segmentation, classification, or cell detection, none of the related works have taken the full calculation of WSI into their experiments.

\subsection{Detection from Whole Slide Image}
The related research extending from local prediction research to WSI calculation is still limited. Gupta et al. \cite{gupta2020region} proposed an automatic ROI identification method for the first time. Although ROI identification can only obtain a rough localization from the WSI, their work considered all the pixels for calculating the WSI, and used all the regions on the WSI for model training, Excellent ROI classification accuracy is achieved on a private dataset containing 10 WSI images, and this study provides an idea of automatically preselecting ROIs for calculating local cancerous cells in WSI.

Cervical cell detection with WSI calculation is also used by several WSI classification researches. Ke et al. \cite{ke2021quantitative} tried to use a nucleus segmentation CNN to guide classification, and then integrated the segmentation results through hand-designed feature engineering. They reported a fabulous WSI classification performance but none of the cell segmentation/classification on external test set. Subsequently, Zhu et al. \cite{zhu2021hybrid} proposed an integrated cervical WSI recognition system, which includes 24 object detection CNNs and other 4 models. To complete the WSI classification task, they also created a new 24 classes to deal with the confusing cervical cancer subclasses, and arranged each object detection CNN for each class. Such a system greatly improves the robustness, showing accuracy and generalization that surpasses pathologists, but it is conceivable that the computing resources required by this system are very large. The author reported its speed is 180 s/WSI. Cao et al. \cite{cao2021novel} designed a new attention module added to the Faster RCNN, a two-stage object detection model, and improved the mAP of cell detection by 2.37\% compared with the baseline. It predicts the confidence of object to weight the features extracted by ResNet to complete WSI classification. None of the performance of cell detection from external/OOD test set is reported. Cheng et al. \cite{cheng2021robust} roughly performed cell detection using low-resolution images, and then performed feature extraction at the location of the object to complete the WSI classification. Wei et al. \cite{wei2021efficient} used a new lightweight object detection model to simultaneously perform cell detection and feature extraction to complete WSI classification.

However, there is rarely a solution and a evaluation when it comes to a complete, cell-level prediction computing all local images on cervical WSI. The training of general deep learning requires the experimental data to conform to the real distribution, but the cervical WSI contains various non-relevant content, that is so-called OOD data, to determine the lesion. Giving such data a full annotation is expensive, scarce, and unnecessary. To develop new method on cell recognition, datasets used in related work are mostly based on local small images generated by pathologists from pre-selected ROIs on WSI. Despite the WSI calculation is completed in WSI classification, their discussion is focus on the classification while the cell detection is just a step of it. Therefore, the related research hardly mentions the generalization and reliability problems that their local prediction model will encounter when calculating the cells of the whole cervical slide, which weakens the contribution of their work.

\section{Method}
\label{sec:method}
This section describes the proposed PolarNet and how this plugin module is working on modern deep learning models. 

\subsection{PolarNet}

\begin{figure}[!t]
\centerline{\includegraphics[width=\columnwidth]{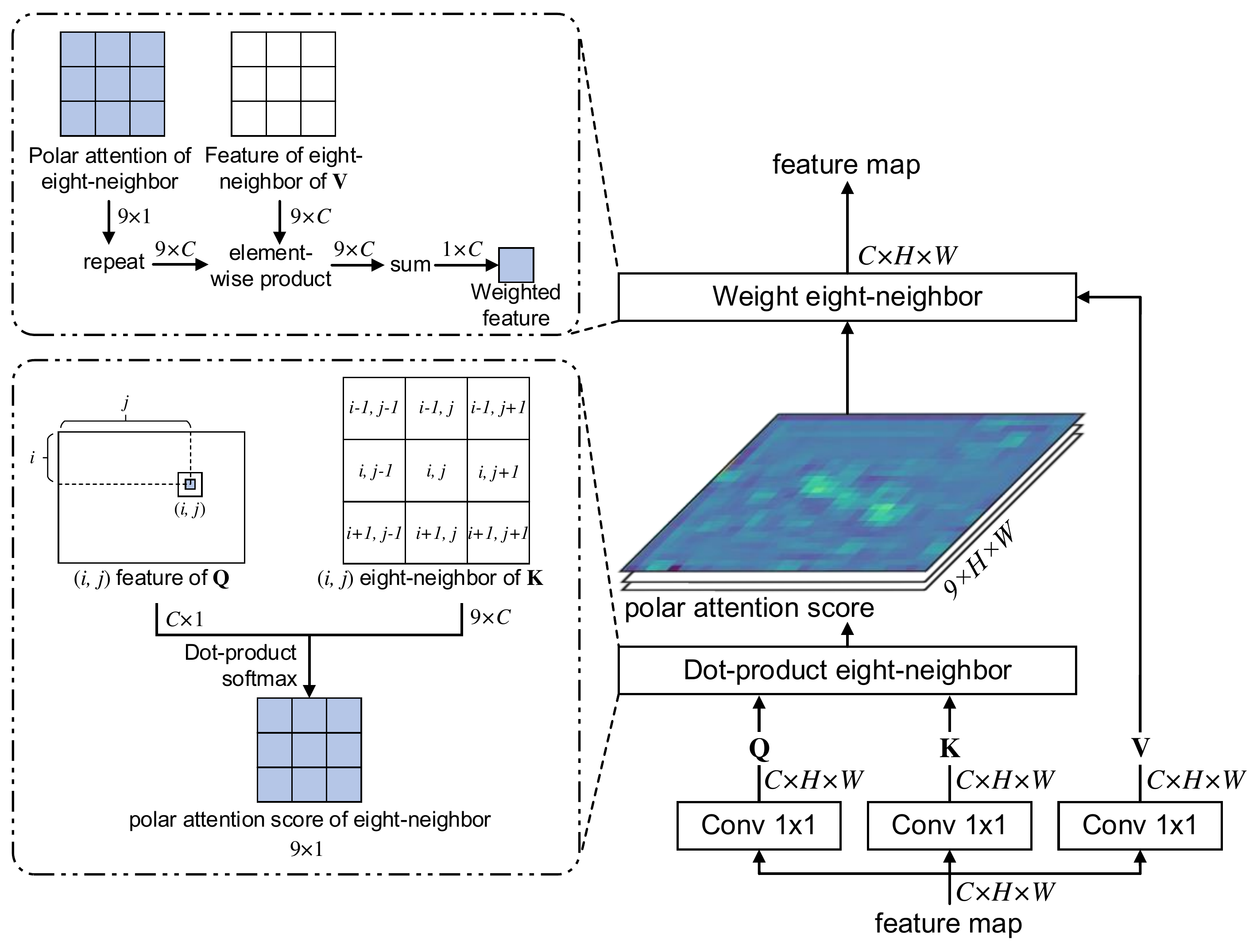}}
\caption{Structure of the proposed PolarNet, where $C$ means the channel number of feature maps. $H$ and $W$ are the height and the width of input feature map, respectively. The dot-product block refers to the Eq. \ref{eq1} and the weighting block refers to the Eq. \ref{eq2}.}
\label{polarnet}
\end{figure}

The PolarNet is designed as shown in Fig. \ref{polarnet}. In the one hand, it obtains a polar attention score matrix by computing the self-attention inside the eight-neighbor of features. On the other hand, it generates a new feature map by weighting with the attention score of different orientations also inside the eight-neighbor area.

The PolarNet is designed to compute the last stage of feature maps from the backbone of a modern model. Thus, we can assume the feature maps are denoted by $\mathbf{x}\in \mathbb{R}^{C\times H\times W}$, the polar attention score matrix by $\mathbf{PAS}\in \mathbb{R}^{9\times H\times W}$, and the output feature maps of the PolarNet by $\mathbf{y}\in\mathbb{R}^{C\times H\times W}$. Then, outputs of PolarNet can be writen by follows:

\begin{equation}
\mathbf{PAS}_{\cdot,i,j}=softmax(
\{\mathbf{Q}_{\cdot,i,j}
\odot
\mathbf{K}^{T}_{\cdot,nei[n]}\}_{n=1,2,\dots,9}
),
\label{eq1}
\end{equation}
\begin{equation}
\mathbf{y}_{c,i,j}=norm(\sum_{n=1}^{9}
\mathbf{PAS}_{n,i,j}
\times
(1+\mathbf{V}_{c,nei[n]})
),
\label{eq2}
\end{equation}
where $i,j$ means the coordinator of $i^{th}$ row and $j^{th}$ column in feature maps, $c$ means the $c^{th}$ channel of feature maps, $\odot$ means the operation of dot-product, $nei\in \mathbb{N}^{9\times2}$ is the index set of eight-neighbor $[(i-1,j-1),(i-1,j),(i-1,j+1),(i,j-1),(i,j),(i,j+1),(i+1,j-1),(i+1,j),(i+1,j+1)]$. $\mathbf Q\in \mathbb{R}^{C\times H\times W}, \mathbf K\in \mathbb{R}^{C\times H\times W}, \mathbf V\in \mathbb{R}^{C\times H\times W}$ means the query, the key, and the value of feature maps, respectively, for computing the self-attention:

\begin{equation}
\mathbf{Q}=Conv^1_{1\times1}(\mathbf x),
\label{eq3}
\end{equation}
\begin{equation}
\mathbf{K}=Conv^2_{1\times1}(\mathbf x),
\label{eq4}
\end{equation}
\begin{equation}
\mathbf{V}=Conv^3_{1\times1}(\mathbf x).
\label{eq5}
\end{equation}

\subsection{Framework of GC Detection}
\label{sec:featscale}

In the general object detection task, multiple models are designed by using different architectures. Such as the anchor-based single-step architecture of YOLO series \cite{redmon2016you}, the multiple-step Faster RCNN \cite{girshick2015fast} and Cascade RCNN \cite{cai2018cascade}, and the anchor-free FCOS \cite{tian2019fcos}. They are both structured by the backbone and the head, where the backbone can employ a neck, like a FPN \cite{lin2017feature} or a PAFPN \cite{liu2018path}, to mix multi-scale feature maps. 

Particularly, whether using the neck in the backbone or not, the PolarNet is plugging into the end part. It computes the last stage of feature maps, the $5^{th}$ stage (downsampling ratio is $2^5=32$), to generate $\mathbf y$ and $\mathbf{PAS}$. The new feature maps $\mathbf y$, then, input to the head of model to predict bounding boxes of GC.

\begin{figure}[!t]
\centerline{\includegraphics[width=\columnwidth]{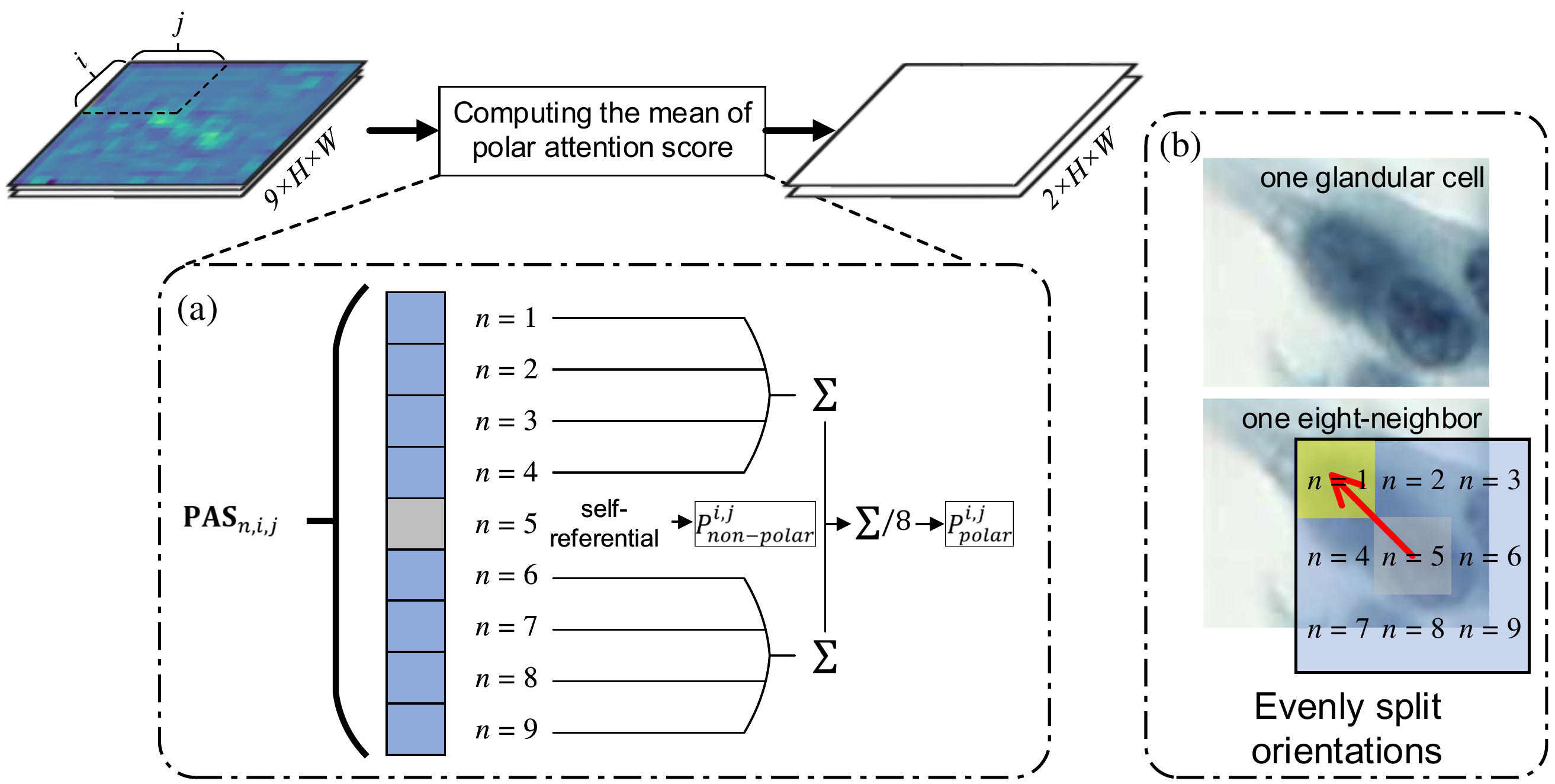}}
\caption{Polar orientations in the eight-neighbor. (a) Using the mean of polar attention score to represent the polar salience of GC. (b) Illustrating orientations in one eight-neighbor of one case of GC. }
\label{polarorientation}
\end{figure}

\begin{figure*}[!t]
\centerline{\includegraphics[width=1.8\columnwidth]{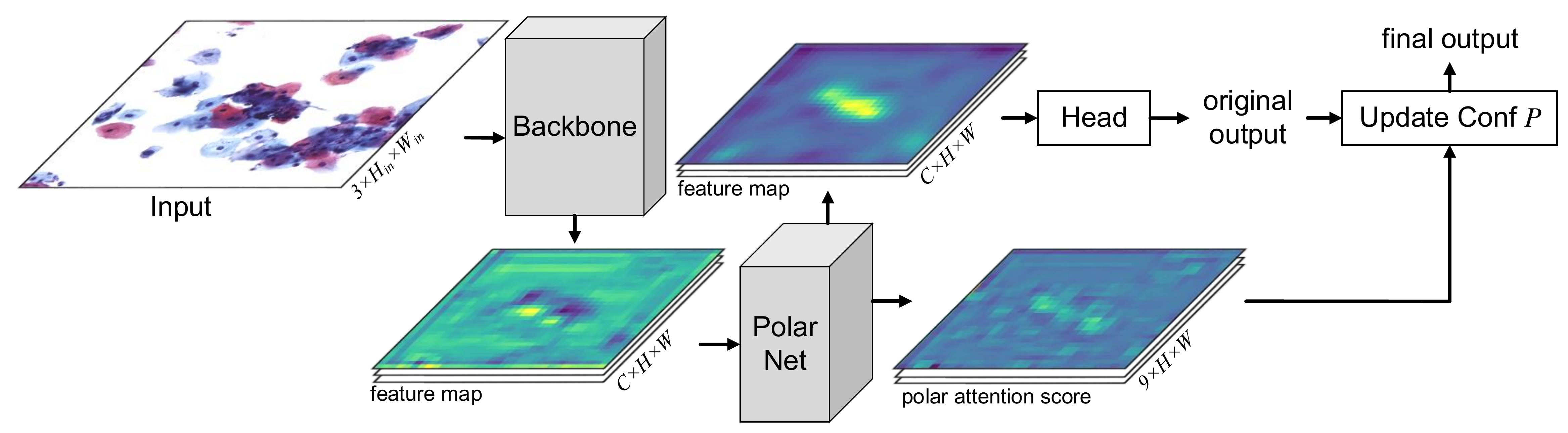}}
\caption{Glandular cell detection framework with PolarNet, where $H_{in}$, $W_{in}$ are the height and the width of input image, respectively.}
\label{framework}
\end{figure*}

The reason of using the last stage of feature maps is pursuing the best fitness between the area size of one feature vector and the physical size of GC. In the PolarNet, polar orientations are split evenly into eight orientations with a self-referential orientation as shown in Fig. \ref{polarorientation}. To properly estimate the true orientation of GC by the PolarNet, one feature vector of input maps should represent the area of around $\frac{1}{3}\sim \frac{2}{3}$ length of GC. Downsampling ratio in the $5^{th}$ stage is just hitting this range. Under the resolution of image in this work, $0.2499\ \mu m/pixel$, one $5^{th}$ stage feature vector represents the $7.9968\ \mu m$ length area, which is $\frac{1}{3}\sim \frac{4}{5}$ times of GC, such as the case in Fig. \ref{polarorientation} (b). Thus, earlier stages, like $4^{th}$ with the ratio $2^4$ or $3^{th}$ with $2^3$, are unsuitable to predict the polar orientation of GC by the self-attention mechanism of eight-neighbor of the PolarNet.

Finally, based on the above knowledge, the GC detection framework can be illustrated as Fig. \ref{framework}. The confidence of GC bounding box is updated by
\begin{equation}
P=(1-\alpha)P_{obj} + \alpha P_{polar},
\label{eq6}
\end{equation}
where $\alpha\in [0,1]$ is the weight of polar salience, and $P_{obj}$ means the objectiveness confidence of original output of model. $P_{polar}$ is the result by converting $\mathbf{PAS}$ matrix to one scalar
\begin{equation}
P_{polar}=\frac{1}{8b_wb_h}\sum_{i=b_x-\frac{b_w}{2}}^{b_x+\frac{b_w}{2}}\sum_{j=b_y-\frac{b_h}{2}}^{b_y+\frac{b_h}{2}}\sum_n^{1\sim4,6\sim9}\mathbf{PAS}_{n,i,j},
\label{eq7}
\end{equation}
where $b_x,b_y,b_w, b_h$ mean the coordinator and the size of one bounding box. It is clear that Eq. \ref{eq7} computes the mean of polar attention score to represent the polar salience of GC as shown in Fig. \ref{polarorientation} (a).

In the training phase, the PolarNet is supervised by cross entropy as similar as the objectiveness loss in modern object detection model
\begin{equation}
L_{PolarNet}=log(NLLLoss([P_{non-polar}\ P_{polar}], P_{gt})),
\label{eq8}
\end{equation}
where $P_{gt}$ means the objectiveness ground truth of one bounding box, and the operation combination of $log$, $NLLLoss$, and $softmax$ in Eq. \ref{eq1} forms the cross entropy loss with
\begin{equation}
P_{non-polar}=\frac{1}{b_wb_h}\sum_{i=b_x-\frac{b_w}{2}}^{b_x+\frac{b_w}{2}}\sum_{j=b_y-\frac{b_h}{2}}^{b_y+\frac{b_h}{2}}\mathbf{PAS}_{5,i,j}.
\label{eq9}
\end{equation}

At this point, any modern model in the object detection task is able to use the prior knowledge, the ubiquitous polarity of GC, to reject false positives of GC from WSI.

\section{Data Preparation}
\label{sec:datapreparation}
This work uses total 486 cervical cytology WSIs from Tongji Union Hospital, Huazhong University of Science and Technology. The scanner used a $20\times$ objective lens with a resolution of $0.2499\ \mu m/pixel$, and used Qupath software \cite{bankhead2017qupath} to complete the local annotation of cervical GC in WSIs.

According to cervical GCs are more sparse than squamous cells, three different sources are set up to make the image datasets for full validation and testing. As shown in Table \ref{tab:data}, the three sources are GC annotation, non-relevant content (NC) annotation and false positive (FP). 

Among them, the GC annotation refers to the area containing glandular cells (clumps) in the cervical slide that is judged to be positive (with the presence of significant atypical GCs), and has two subclasses: AGC (Atypical Glandular Cells) and nGEC (normal Glandular Epithelial Cells).
Non-relevant content (NC) annotation refers to some small areas from negative WSIs (no significant atypical glandular cells are present) that do not contain any GC. 
False positive (FP) refers to the wrong predictions in the test set of GC annotation source by a modern model, YOLOX-l \cite{ge2021yolox}. The source of the first GC annotation is provided by the pathologist, and the latter two are generated by authors. All annotations have been reviewed by the pathologist for the double check. 

\begin{table}
\centering
\caption{Data set details.}
\label{tab:data}
\setlength{\tabcolsep}{3pt}
\begin{tabular}{lccccc}
\hline
source & & slide num. & image num. & AGC num. & nGEC num. \\
\hline
\multirow{3}*{GC ann.} & train & 221 & 8274 & 8759 & 17190 \\
                       & val & 66 & 1032 & 258 & 1331\\
                       & test & 44 & 779 & 226 & 1364 \\
NC ann. & test & 45 & 1280 & 0 & 0 \\
FP & test & 110 & 3496 & 0 & 0 \\
\hline
total & & 486 & 14861 & 9243 & 19885\\
\hline
\multicolumn{6}{p{210pt}}{Note: num. means number, ann. means annotation, NC means non-relevant content, FP means false positive by a modern model.}

\end{tabular}
\end{table}

Images listed above are used with the same size, $1024\times1024$, and the same original resolution. 
\begin{itemize}
    \item While the images are from GC annotations, first, those GC annotations on a slide that are close enough (smaller than the size of image) are used as a set. Then, a larger cell image is cropped from WSI centered on such set with the size of $1536 \times 1536$ and saved in the hard disc. During training, $1024 \times 1024$ regions are randomly cropped from the larger images to ensure the diversity of learning samples, and $1024 \times 1024$ are cropped from the center for validation and testing.
    \item While the images are from non-relevant content annotation, $1024 \times 1024$ cell images are cropped from WSI, randomly, for testing.
    \item The false positive source is obtained by a YOLOX-l \cite{ge2021yolox} model that trained with the first source, GC annotations, in our data. It is from external positive WSIs ($n=110$) that are entirely inferred by the well-trained YOLOX-l. The top bounding boxes ranking by the confidence (named as top-$N$ results) of every WSI are visualized using the Qupath tool, and manually judged one by one. Results that are not GCs are then cropped in the same way of above sources for testing. During the manual judgement, all top-100 in the first 30 slides and 20 random results of top-100 in the last 80 slides are reviewed.
\end{itemize}

\begin{figure}[!t]
\centerline{\includegraphics[width=\columnwidth]{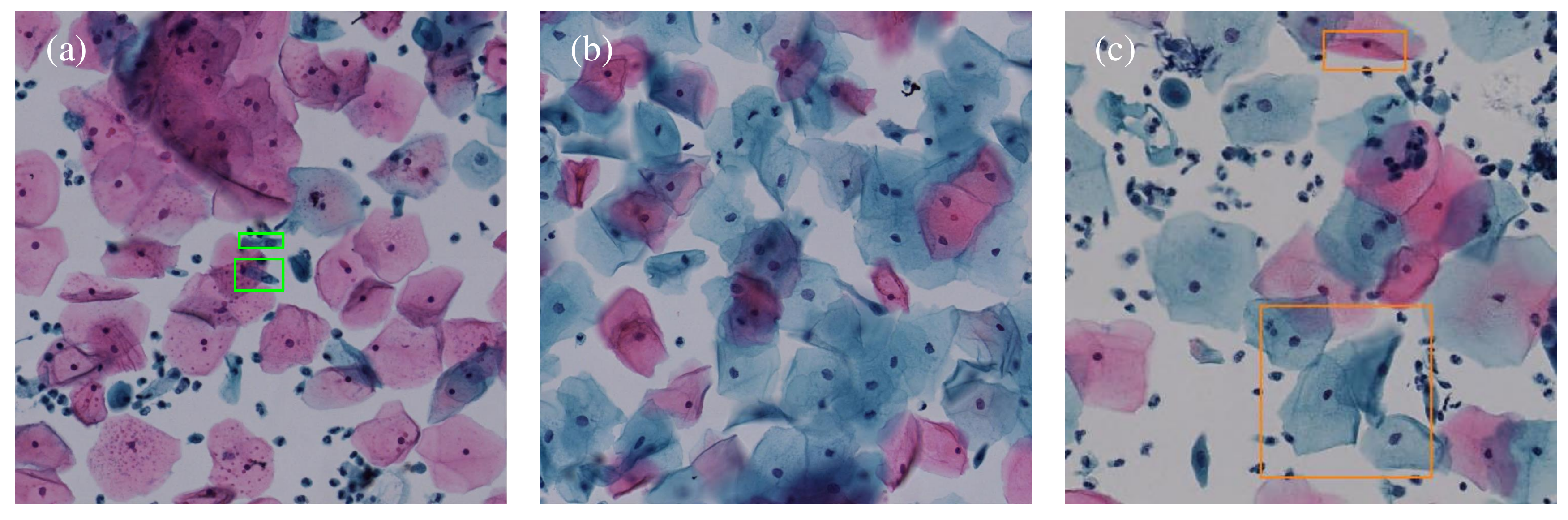}}
\caption{Examples of GC used in this work. (a) Image with nGEC annotations (green boxes). (b) Image containing non-relevant content. (c) Image with false positives (orange boxes) by the well-trained YOLOX-l.}
\label{dataexample}
\end{figure}

\begin{figure*}[!t]
\centerline{\includegraphics[width=2\columnwidth]{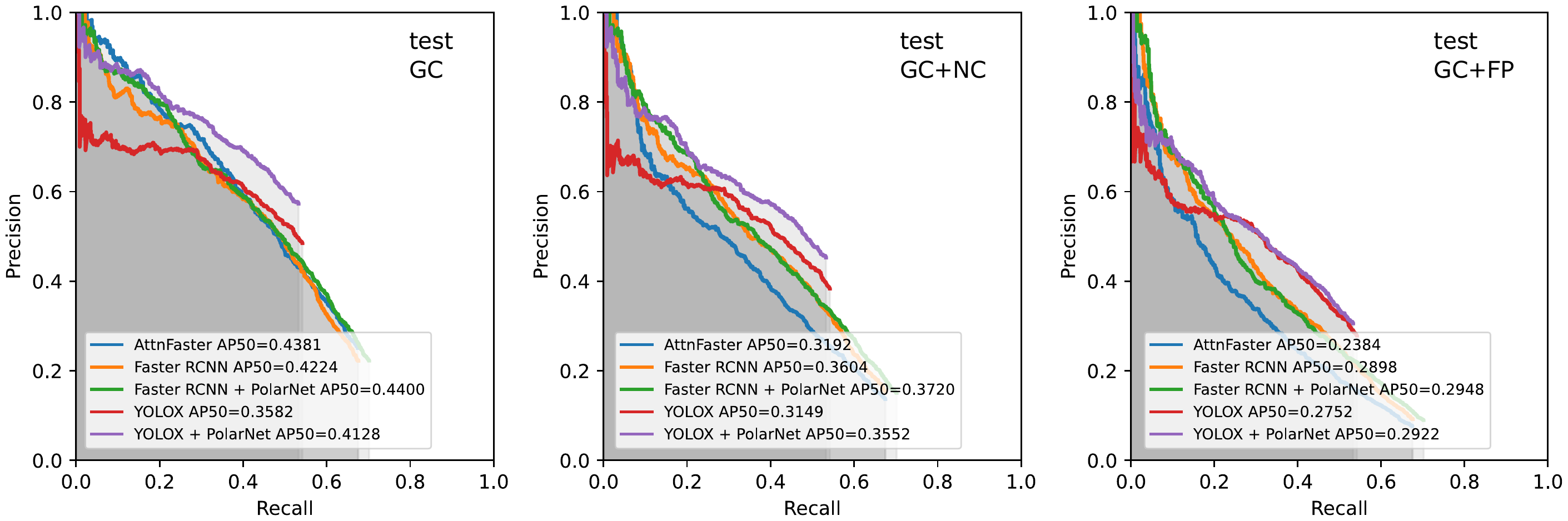}}
\caption{Precision-recall curves on three test data setting.}
\label{pr_curve}
\end{figure*}

Among the results of above manual judgement, it is interesting that the average accuracy of first 30 is only 0.1445. Although this model is very popular in the industry, the accuracy of only 2 slides exceeded 0.9 due to the fact that the cervical GC's prior knowledge is not taken into the model. It is worth noting that this phenomenon of external testing on YOLOX-l indicates that standard experimental results, which shown an average precision of more than 0.7 on the validation set, are unreliable when facing the sparse GC from real WSI. Some examples are shown in Fig. \ref{dataexample}.

It is worth mentioning that testing with the non-relevant and false sources, the images without the presence of GCs accounted for 62.17\% and 81.78\% of the total, respectively. Therefore, the unreliability of the model can be fully demonstrated in both. In fact, the proportion of images without GCs in the whole slide generally exceeds 90\%, so testing with those two sources can simulate the performance of deep learning models for GC detection from WSI.

\section{Experiments}
\label{sec:exp}
\subsection{Evaluation Criteria}

First of all, for the GC detection from $1024\times 1024$ image, this work uses the common evaluation criteria, AP50 (Average Precision with 50\% IoU threshold), and its calculation formula is
\begin{equation}
\text{Recall} = \frac{TP}{TP+FN}, \text{Precision} = \frac{TP}{TP+FP},
\label{eq10}
\end{equation}
where $TP$, $FP$, and $FN$ are true positive, false positive, and false negative, respectively.
Since AP50 is obtained by calculating the area under the Precision-Recall curve, we also provide the curve to show the GC detection performance of models in detail.

Second, for the task of GC detection from cervical cytology WSI, this work uses the top-$N$ accuracy of GC to demonstrate the reliability of models. It is calculated by $\frac{TP}{2(TP+FP)} + \frac{TN}{2(TN+FN)}$ with $TN$ the true negative.

\subsection{Training Environments}

All the experiments in this work are implemented with the PyTorch deep learning library \cite{paszke2019pytorch} on a Win10 OS computer. Model training is performed using the SGD optimizer \cite{bottou2010large}, and the learning rate descent strategy is the common stepwise, from $5 \times 10^{-3}$ to $5 \times 10^{-6}$, learned at epochs 25, 50, and 80, respectively. The rate decay is 0.1, and the maximum training epoch is 100. Memory is 128 Gb, CPU is a Xeon® 6134 @3.20 GHz, GPU is a Tesla P40. The comparison methods are trained using the code published by the relevant research, and the general popular model uses the implementation version of TorchVision model zoo. When using the proposed PolarNet, only one hyper-parameter $\alpha$ is always set as 0.5.

\subsection{Comparing Methods}

Regarding the comparison methods, since YOLOX and Faster RCNN are representative single-stage and double-stage methods commonly used in general tasks. AttnFaster \cite{cao2021novel}, that is specially designed for cervical cytology based on attention mechanism, is also chosen to test. In addition, more existing models are compared: one anchor-free model FCOS, and one multi-stage model Cascade RCNN \cite{cai2018cascade}.


\subsection{GC Detection Results}

As mentioned in Section \ref{sec:datapreparation}, the experimental data in this work were produced by three different sources, each of which represented different levels of sparsity and difficulty of GC detection, to verify the effectiveness of models on real distribution of data from WSI. For fully comparing the different testing scenarios, this section first shows the test results on all three sources of data, and then shows the detailed test results of GC, which has two subclasses.

\begin{table}
\centering
\caption{GC Detection AP50.}
\setlength{\tabcolsep}{3pt}
\begin{tabular}{lccccl}
\hline
\multirow{2}*{Model Name} & 
val &
\multicolumn{3}{c}{test} &
\multirow{2}*{average of test}\\
& GC & GC & GC+NC & GC+FP & 
\\
\hline
AttnFaster + R50   & 0.570 &0.438 & 0.319 & 0.238 & 0.332\\
\hline
FCOS + R50 & 0.323 & 0.272 & 0.183 &0.112 & 0.189\\
$\sim$ w/ PolarNet  & 0.354 & 0.312 & 0.181 & 0.096 & 0.196 (\textcolor{green}{+0.007})\\
YOLOX-m & 0.592 & 0.358 & 0.315 & 0.275 & 0.316\\
$\sim$ w/ PolarNet  & 0.528 & 0.413 & 0.355 & 0.292 & 0.353 (\textcolor{green}{+0.037})\\
Faster RCNN + R50   & 0.499 & 0.422 & 0.360 & 0.290 & 0.357\\
$\sim$ w/ PolarNet  & 0.513 & 0.440 & 0.372 & 0.295 & \textbf{0.369} (\textcolor{green}{+0.012})\\
Cascade RCNN + R50   & 0.437 & 0.294 & 0.264& 0.208 & 0.255\\
$\sim$ w/ PolarNet  & 0.444 & 0.328 & 0.271& 0.210 & 0.270 (\textcolor{green}{+0.015})\\
\hline
\end{tabular}
\label{tab:det-res}
\end{table}

\subsubsection{Main results}
The GC detection results from small images are shown in Table \ref{tab:det-res}. It is obvious that all SOTA models with PolarNet appear the improvement of AP50 comparing to their original version.
That can show the effectiveness of PolarNet. And, Faster RCNN with PolarNet always shows the highest AP50 on every test setting and the average. YOLOX with PolarNet is able to exceed AttnFaster in average by 0.021, while the original YOLOX is 0.016 lower than AttnFaster.
Interestingly, comparing the results between three different test settings, the AP50 of only GC source can maintain a similar level of AP50 as the validation set. But after adding images without target presence (non-relevant or false positive source), AP50 appears some relatively significant drops. Even for the most complex Cascade RCNN, which performs best on the general dataset, the AP50 of test set GC+FP is only 0.255.
Due to the lower AP50 scores of FCOS and Cascade RCNN among the general models, we use the more powerful and more general YOLOX and Faster RCNN for subsequent experiments.

\subsubsection{Precision-recall curves}
The precision-recall (P-R) curves of above models are shown in Fig. \ref{pr_curve}. It can be seen that the green of PolarNet used in the P-R curve of the Faster RCNN model has higher precision than orange and blue, and the red P-R curve of YOLOX is also lower in precision than adding PolarNet. YOLOX. For Recall, P-R curves for the same model infrastructure all exhibit similar maximum recall. While two YOLOXs' recalls are significantly lower than Faster RCNN, their precisions are higher when curves start. The relatively low precision at the beginning of the curve means it will be less effective when encountering external test sets. This is because the curve is drawn with the results sorted by confidence, and hence the results with high confidence are less accurate when that happened. It can be seen that YOLOX after adding PolarNet is improved at the beginning of the curve.

\subsubsection{Detailed results}
In object detection task, the calculation of AP criteria depends on the threshold of the intersection over union between bounding boxes of predictions and annotations. For example, AP50 used above needs to be greater than 50\% to be judged as $TP$. Obviously, a larger threshold can reflect the accurate size and position of predictions. As shown in Table. \ref{tab:detailed-res}, in order to demonstrate the accuracy of sizing and positioning of predictions, we list the results in AP60 and AP70. It can be seen that, similar to the trend in Table. \ref{tab:det-res}, AP60 and AP70 of Faster RCNN with PolarNet still both perform best on test set of the GC+FP source, while YOLOX with PolarNet ranks third in AP60 and AP70 . Obviously, compared to AP50, all models show a small drop at AP60 and a significant drop at AP70. This shows that the accurate detection of cervical GC is still a challenging problem. Furthermore, AP50 of two subclasses are also shown in Table. \ref{tab:detailed-res}. The two models using PolarNet performed the second with 0.112 and the first with 0.336 in AGC and nGEC, respectively. And, AP50 of AGC is generally lower than that of nGEC, because the number of AGC annotations in the training data is less than half of that of nGEC.

\begin{table}[!t]
\centering
\caption{Detailed results of GC detection.}
\label{tab:detailed-res}
\setlength{\tabcolsep}{3pt}
\begin{tabular}{lccccc}
\hline
Model Name &
nGEC &
AGC &
AP50 &
AP60 &
AP70\\
\hline
AttnFaster & 0.267 & 0.079 & 0.238 & 0.168 & 0.077\\
Faster RCNN & 0.328 & 0.075 & 0.290 & 0.207 & 0.091\\
$\sim$ w/ PolarNet & \textbf{0.336} & 0.083 & \textbf{0.295} & \textbf{0.210} & \textbf{0.093}\\
YOLOX-m & 0.316 & \textbf{0.184} & 0.275 & 0.184 & 0.080\\
$\sim$ w/ PolarNet & 0.325 & 0.112 & 0.292 & 0.190 & 0.078\\
\hline
\multicolumn{6}{p{170pt}}{Note: results of subclasses nGEC and AGC are both AP50.}
\end{tabular}
\end{table}

\begin{table}[!h]
\centering
\caption{WSI GC Detection Top-N Accuracy in Average.}
\label{tab:wsi-res}
\setlength{\tabcolsep}{3pt}
\begin{tabular}{lcccc}
\hline
Model Name &
$N_{pred}$ &
Top-5 &
Top-10 &
Top-20\\
\hline
AttnFaster & 20.00 & 0.281$\pm$0.311 & 0.284$\pm$0.274 & 0.271$\pm$0.241 \\
Faster RCNN & 20.00 & 0.424$\pm$0.344 & 0.410$\pm$0.303 & 0.355$\pm$0.261 \\
$\sim$ w/ PolarNet & 20.00 & \textbf{0.429}$\pm$0.364 & \textbf{0.411}$\pm$0.344 & \textbf{0.386}$\pm$0.282 \\
YOLOX-m & 18.89 & 0.117$\pm$0.236 & 0.0.108$\pm$0.212 & 0.122$\pm$0.204 \\
$\sim$ w/ PolarNet & 20.00 & 0.190$\pm$0.283 & 0.205$\pm$0.267 & 0.126$\pm$0.149 \\
\hline
\multicolumn{5}{p{230pt}}{Note: $N_{pred}$ means the average number of predictions in every WSI, $\pm$ followed by the standard deviation of accuracies of all WSIs.}
\end{tabular}
\end{table}

\subsection{WSI GC Detection Top-N Results}
When the cervical GC detection model is practically applied to cervical cancer CAD, the top-$N$ results can guide pathologists to prioritize the screening of suspicious lesions. Therefore, the top-$N$ results of models on those external slides ($n=110$) can demonstrate its performance from the perspective of cervical cancer auxiliary diagnosis on WSI. This section shows the accuracy of top-$N$ results, with higher accuracy indicating greater potential for the model to be applied to CAD.

\begin{figure}[!b]
\centerline{\includegraphics[width=0.9\columnwidth]{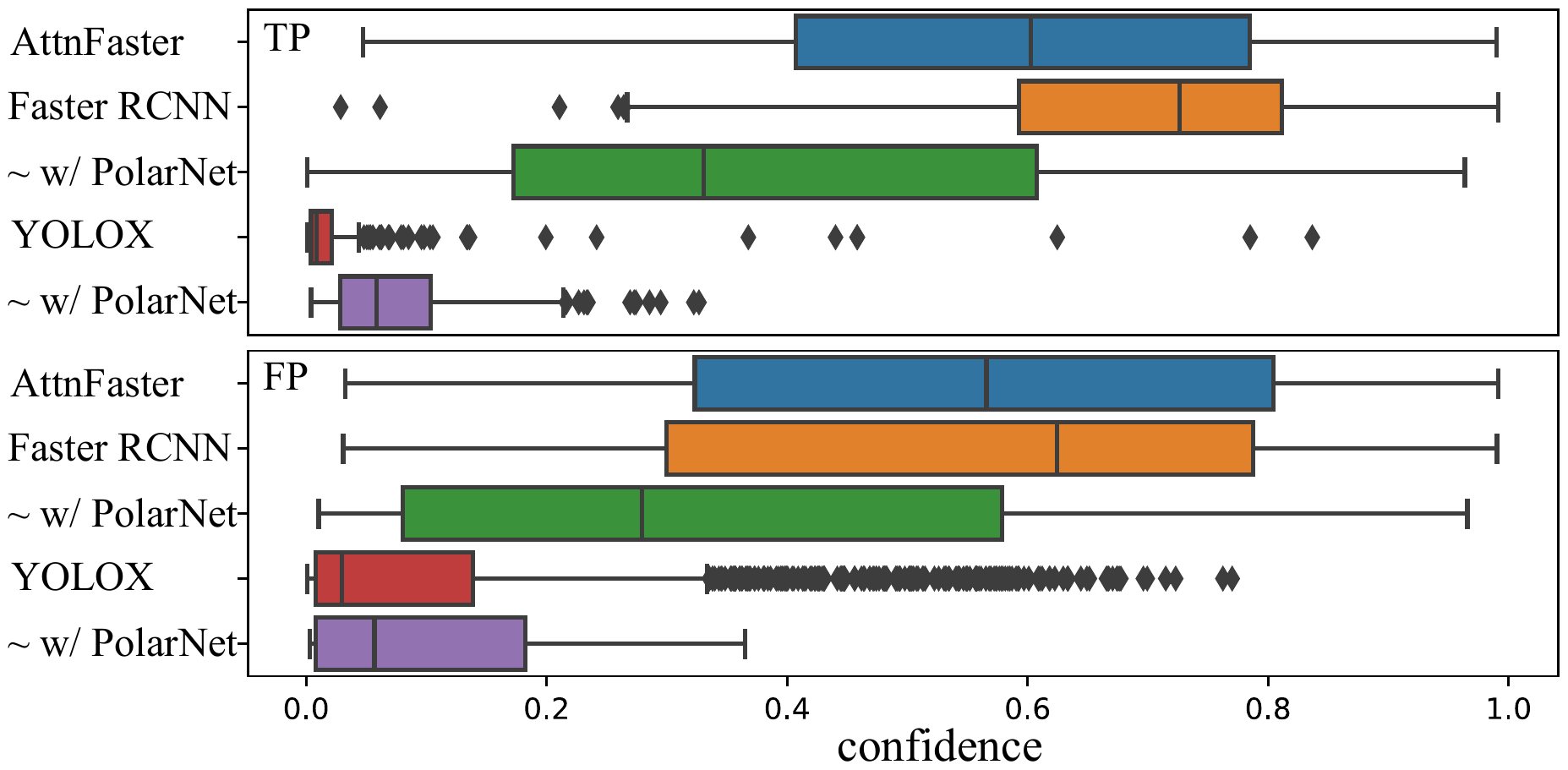}}
\caption{The box plot showing the distribution of WSI GC detection top-20 results.}
\label{wsi-detconf}
\end{figure}

As shown in Table. \ref{tab:wsi-res}, the accuracy of top-$N$ results when $N=5$, $10$, and $20$ are listed respectively. It is clear that Faster RCNN with PolarNet performs the best among the three different $N$. As $N$ decreases, its accuracy increases from 0.386 to 0.429. This shows that it has a reasonable confidence distribution, such as higher confidence results have higher accuracy. In fact, Faster RCNN is the only remaining model with the same trend, and the other remaining three models show a drop in accuracy when using higher confidence results. In YOLOX models, its accuracy is much lower than the other models as shown in Fig. \ref{wsi-detconf}. This situation can also be reflected from $N_{pred}$, the average number of predictions. There are some slides that cannot give the complete 20 detection results, even though YOLOX used a very low confidence threshold (0.0001). Although adding PolarNet can alleviate this situation, the 
top-$N$ accuracy of YOLOX with PolarNet is still at least about 0.08 lower than the double-stage Faster RCNN. This shows that the double-stage model is the most reliable when faced with real external WSI comparing with others.

\subsection{Ablation Studies}
In order to further reveal the role of PolarNet in object detection models and prove the effectiveness of polar attention for cervical GC detection, two ablation studies are performed in this section:
1) The contribution of $P_{polar}$, the mean of polar attention score, in the confidence $P$ updating Eq. \ref{eq6}.
2) The effect of feature scale (see Section \ref{sec:featscale}) on the estimation of GC polarity in PolarNet.

\subsubsection{The contribution of $P_{polar}$}
Based on Eq. \ref{eq6} that changing the value of $\alpha$ can change the contribution of $P_{polar}$. Especially, when $\alpha = 0$, $P_{polar}$ contributes nothing in the role of PolarNet but only the new feature maps guided by polar attention. Therefore, observation of the $\alpha$-AP50 curve can simultaneously demonstrate the effectiveness of $P_{polar}$ and polar attention matrix. 

\begin{figure}[!t]
\centerline{\includegraphics[width=\columnwidth]{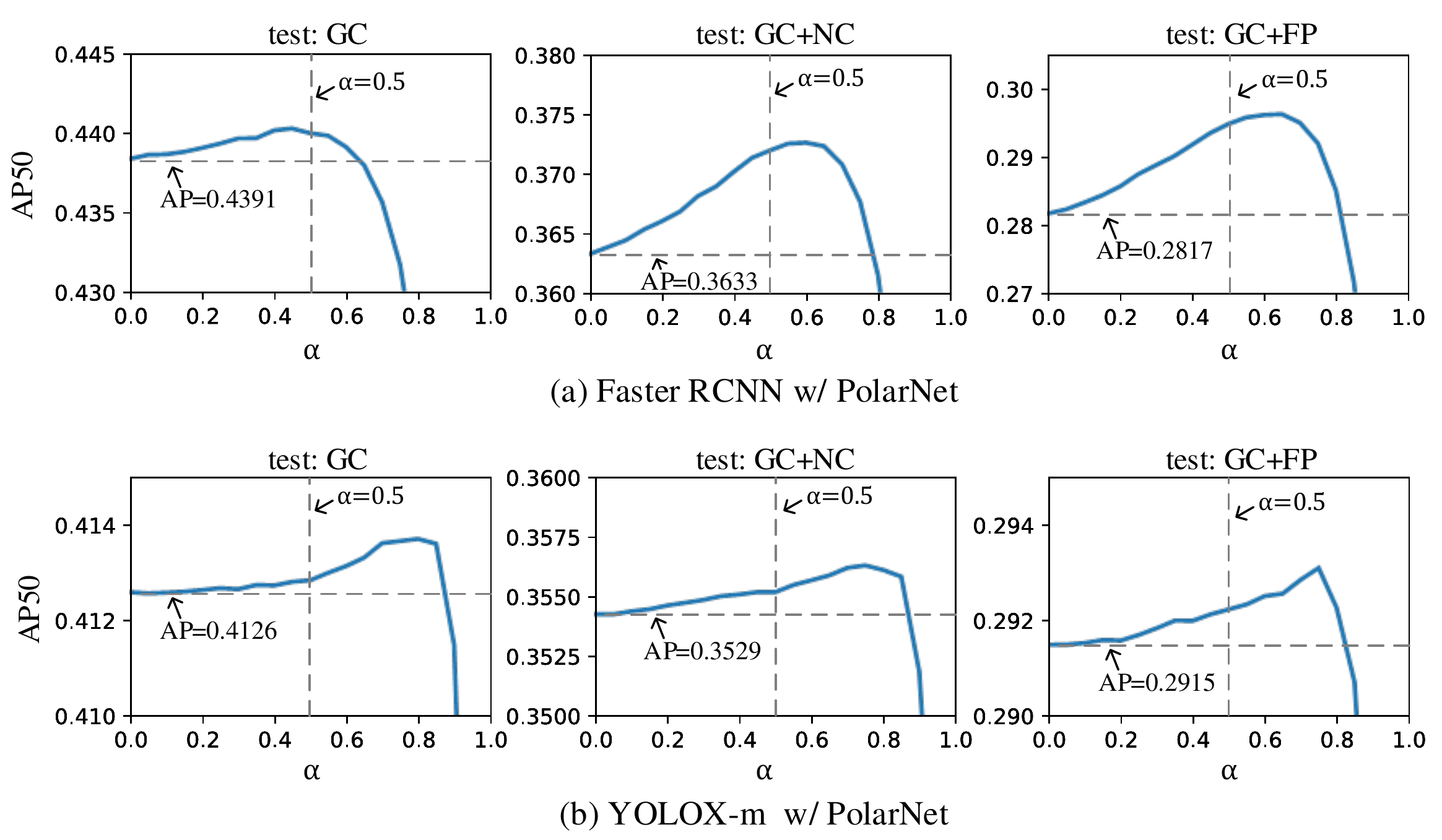}}
\caption{$\alpha$-AP50 curves to show the significance of $P_{polar}$, the mean value of polar attention score.}
\label{ablation-alpha}
\end{figure}

Fig. \ref{ablation-alpha} shows the $\alpha$-AP50 curve by two types of model. It can be seen that whether it is a single-stage YOLOX or a multi-stage Faster RCNN, the AP50 raises when $\alpha$ increasing, and reaches a peak after the default setting $\alpha = 0.5$ in this work. This indicates that the polar score provides a positive contribution to the GC detection results. Moreover, it can be seen that when $\alpha = 0$, the AP50 of the model is still superior to the results of other models in Table. \ref{tab:det-res} on the three test settings. This shows that polar attention also provides an important contribution to the feature representation of the model.

\begin{figure*}[!b]
\centerline{\includegraphics[width=2\columnwidth]{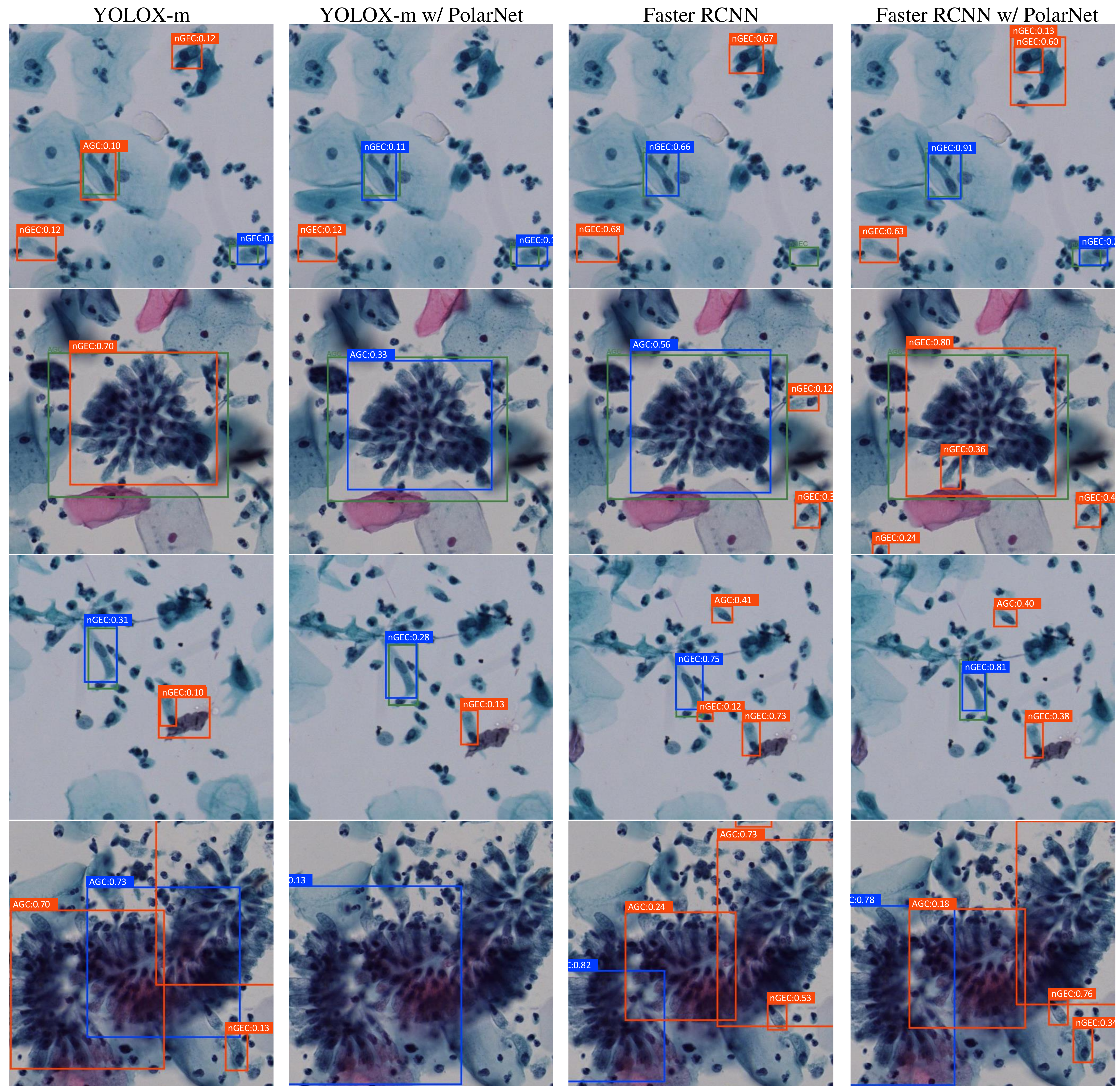}}
\caption{GC detection examples by four models to show the ability of rejection of false GCs after using PolarNet, where green boxes are ground truth, oranges are false positives, and blues are true positives.}
\label{detexample}
\end{figure*}

\subsubsection{The effect of feature scale}
As described in Section. \ref{sec:featscale}, since the physical size of a single GC is relatively fixed, PolarNet calculates its polarity orientation by feature maps that are sensitive to the scale of GC. This part verifies that the scale, $2^5$, in $5^{th}$ stage that covers about $\frac{1}{3}\sim\frac{4}{5}$ of the cell size obtains optimal polar attention. We test all available feature scales for both the single-stage model and the double-stage model, showing the effect of feature scale on PolarNet.

As shown in Table. \ref{tab:ablation-res}, the results AP50 of cervical GC detection using different scales of feature for PolarNet. Obviously, the model using the largest scale, $2^5$, performs the best on all test settings. This shows that the theoretical foundation for computing polar attention is valid. In addition, it can be seen that after PolarNet uses smaller scales, most of the AP50, such as non-bold and non-underlined, are even lower than the corresponding original model shown in Table. \ref{tab:det-res}. It shows that when PolarNet uses an inappropriate scale, the polarity calculation via self-attention mechanism in eight-neighbor no longer has a positive effect.

\begin{table}[!t]
\centering
\caption{GC Detection AP50 on Test Set using Different Feature Scales.}
\label{tab:ablation-res}
\setlength{\tabcolsep}{6pt}
\begin{tabular}{p{7pt}cccccc}
\hline
\multirow{2}*{Scales} &
\multicolumn{3}{c}{Faster RCNN w/ PolarNet} &
\multicolumn{3}{c}{YOLOX-m w/ PolarNet} \\
& GC & GC+NC & GC+FP & GC & GC+NC & GC+FP \\
\hline
$2^5$ & \textbf{0.443} & \textbf{0.376} & \textbf{0.295} & \underline{0.413} & \textbf{0.355} & \textbf{0.292} \\
$2^4$ & 0.382 & 0.321 & \underline{0.256} & \textbf{0.436} & \underline{0.324} & \underline{0.222} \\
$2^3$ & 0.396 & 0.320 & 0.228 & 0.343 & 0.266 & 0.201 \\
$2^2$ & \underline{0.444} & \underline{0.338} & 0.247 & - & - & - \\
\hline
\end{tabular}
\end{table}

\subsection{Qualitative Results}

Finally, Fig. \ref{detexample} shows the detection examples of cervical GC for the four models for comparison, including the detection examples of two groups of independent GCs and another two of GC clusters. First, in all examples by every model, it is difficult to avoid giving false GCs (orange box), which is caused by the sparseness of GC in WSI. Then, compared with Faster RCNN and YOLOX-m, pulgging PolarNet to modern models can generally reduce the confidence of false bounding boxes of GC. In these four sets of examples, except for the Faster RCNN in the second row, all the others have removed false positives (orange boxes) by PolarNet's scoring of polar attention, and in some cases even corrected false negatives, such as the first and second lines, where new blue box is appearing. These examples visually illustrate the role of PolarNet in GC detection.

\subsection{Computational Costs}
In additional, this work also complete the C++ deployment of the trained cervical GC detection model.\footnote{C++ program is released in  \href{https://github.com/Chrisa142857/You-Only-Look-Cytopathology-Once/tree/main/cpp}{https://github.com/Chrisa142857/You-Only-Look-Cytopathology-Once/tree/main/cpp}} The C++ program is able to uses any modern model to complete the cervical GC detection from WSI. The only need is to change the model path and set some hyper-parameters in the command line. To test computational costs,  this section involves 5 object detection models that tested in above experiments: AttnFaster, Faster RCNN, Faster RCNN  with PolarNet, YOLOX-m, and YOLOX-m with PolarNet.

Table. \ref{tab:compcost} shows computational costs using the proposed PolarNet in our C++ program in detecting GCs from WSIs. Although the YOLOX model can achieve the minimum average time of 107.6 s, its accuracy drops severely, only 12.2\%. The best model Faster RCNN with PolarNet improves the accuracy by 3.1\% compared with its original version by sacrificing the time-consuming of 14.4 s in average.

\begin{table}[!t]
\centering
\caption{Computational Costs of GC Detection from WSIs ($n=110$).}
\label{tab:compcost}
\setlength{\tabcolsep}{7.5pt}
\begin{tabular}{p{50pt}lll}
\hline
\makecell[c]{Model Name} &
\makecell[c]{Total Time} &
\makecell[c]{Average Time} &
\makecell[c]{Top-20 Acc.} \\
\hline
AttnFaster & 6.8 hr & 221.3 s & 27.1\%\\
Faster RCNN & 7.5 hr & 244.3 s & 35.5\%\\
$\sim$ w/ PolarNet & 7.9 hr (\textcolor{red}{+0.4}) & 258.7 s (\textcolor{red}{+14.4}) & \textbf{38.6\%} (\textcolor{green}{+3.1})\\
YOLOX-m & \textbf{3.3 hr} & \textbf{107.6 s} & 12.2\%\\
$\sim$ w/ PolarNet & 4.6 hr (\textcolor{red}{+1.3}) & 151.1 s (\textcolor{red}{+43.5}) & 12.6\% (\textcolor{green}{+0.4})\\
\hline
\end{tabular}
\end{table}

\section{Conclusion}
\label{sec:conclusion}
Current cervical cell detection works generally do not consider the morphology and sparseness of GC in WSI. But the slender shape of GC is easily confused, which makes the general model have serious false positives when detecting WSI. In order to improve the reliability of GC detection from WSI, a PolarNet is designed in this paper. It evaluates the polarity of cells by calculating eight-neighbor self-attention, and generates polarity score and polarity attention to guide confidences and feature maps, respectively, to eliminate pseudo-GC with insignificant polarity. The experimental results show that PolarNet can improve the Top-20 accuracy of GC detection from WSI by 8.8\% on the presence of sacrificing the computational time per slide of 14.4 s.

\section{Discussion}
\label{sec:discuss}
In this paper, a novel network PolarNet is proposed, which can effectively eliminate pseudo-GCs with insignificant polar orientations in the detection results, and more reliably complete GC detection from WSI. The network obtains the polar average score in the detection frame by calculating the eight-neighborhood attention score in the feature map to judge whether the polarity direction of the cell is significant, and can be added to any general model as a module. In order to obtain suitable polarity orientations, it uses the appropriate scale of the $5^{th}$ stage feature map, so that the polarity orientations can be obtained from general object detection model to cover $\frac{1}{3} \sim \frac{4}{5}$ shape of a single GC. In experimental sections, in order to demonstrate the detection performance from WSI, PolarNet was first tested on three small images with different degrees of sparseness of GC and showed superior average accuracy, and then top-$N$ accuracy was also shown the cervical GC detection from WSI. In addition, two ablation studies further demonstrate the positive effect of polar attention score and the new polar guided feature maps of PolarNet. In general, the PolarNet designed based on the prior knowledge of cervical GC can effectively improve the reliability of GC detection from WSI with OOD data.

However, there is still room to improve the proposal. The current PolarNet obtains the polar attention score by calculating the feature map at the fixed scale, which is too sensitive to the change of scale as shown in the second ablation study. Although the scale of a single cervical WSI is hardly to change, the network suffers when encountering data from multiple sources. Therefore, it is very important to further develop scale-invariant PolarNet.

Since the deployment of C++ program is working nice, further clinical experiments could be carried out to verify the effectiveness in real-world tasks. The proposed program has a good improvement with a slight time cost under current computational resources, and its usefulness and weakness can be further revealed by running to the actual cervical cancer CAD.

\section*{Acknowledgment}

The authors want to thank pathologists and organizations that
provided the raw data and the manual annotations. As well
as the Collaborative Innovation Center for Biomedical Engineering and the Britton Chance Center and MOE Key Laboratory for Biomedical Photonics should be greatly appreciated
for their platforms and devices. This work is supported by the National Natural Science Foundation of China (NSFC) projects (grant 61721092), China Postdoctoral Science Foundation (grant 2021M701320) and the director fund of the Wuhan National Laboratory for Optoelectronics
(WNLO).

\bibliographystyle{IEEEtran}
\bibliography{tmi}

\end{document}